\documentclass{article}
\usepackage{bm}
\usepackage{xcolor}
\usepackage{amsthm}
\usepackage{amsmath}
\usepackage{amssymb}
\usepackage{multirow}
\usepackage{enumitem}
\usepackage{makecell}
\usepackage{booktabs}
\usepackage{hyperref}
\usepackage{graphicx}
\usepackage{microtype}
\usepackage{subfigure}
\usepackage{mathtools}
\usepackage{threeparttable}
\usepackage[page]{appendix}
\usepackage[textsize=tiny]{todonotes}
\usepackage[capitalize,noabbrev]{cleveref}

\newcommand{\RN}[1]{\textup{\uppercase\expandafter{\romannumeral#1}}}
\usepackage[accepted]{icml2024}

\icmltitlerunning{DynamicLight: Two-Stage Dynamic Traffic Signal Timing}
\begin{document}
\twocolumn[
\icmltitle{DynamicLight: Two-Stage Dynamic Traffic Signal Timing}
\icmlsetsymbol{equal}{*}

\begin{icmlauthorlist}
\icmlauthor{Liang Zhang}{lzu}
\icmlauthor{Yutong Zhang}{bupt}
\icmlauthor{Shubin Xie}{lzu}
\icmlauthor{Jianming Deng}{lzu}
\icmlauthor{Chen Li}{ngy}
\end{icmlauthorlist}

\icmlaffiliation{lzu}{State Key Laboratory of Herbage Improvement and Grassland Agro-ecosystems, College of Ecology, Lanzhou University, Lanzhou 730000, China }
\icmlaffiliation{bupt}{School of Artificial Intelligence, Beijing University of Posts and Telecommunications}
\icmlaffiliation{ngy}{Graduate School of Informatics, Nagoya University, Chikusa, Nagoya 464-8601, Japan}

\icmlcorrespondingauthor{Jianming Deng}{dengjm@lzu.edu.cn}

\icmlkeywords{traffic signal control, reinforcement learning, two-stage control}

\vskip 0.3in
]
\printAffiliationsAndNotice{\icmlEqualContribution}

\begin{abstract}
Reinforcement learning (RL) is gaining popularity as an effective approach for traffic signal control (TSC) and is increasingly applied in this domain. However, most existing RL methodologies are confined to a single-stage TSC framework, primarily focusing on selecting an appropriate traffic signal phase at fixed action intervals, leading to inflexible and less adaptable phase durations. To address such limitations, we introduce a novel two-stage TSC framework named \textit{DynamicLight}. This framework initiates with a phase control strategy responsible for determining the optimal traffic phase, followed by a duration control strategy tasked with determining the corresponding phase duration. Experimental results show that DynamicLight outperforms state-of-the-art TSC models and exhibits exceptional model generalization capabilities. Additionally, the robustness and potential for real-world implementation of DynamicLight are further demonstrated and validated through various DynamicLight variants. The code is released at \url{https://github.com/LiangZhang1996/DynamicLight}.
\end{abstract}

\section{Introduction}
\label{sec:intro}
Signalized intersections dominate as the primary type of road junctions in urban landscapes, where traffic signal control (TSC) plays a pivotal role in ensuring effective traffic management. Established methods, exemplified by FixedTime~\cite{fixedtime}, GreenWave~\cite{greenwave}, SCATS~\cite{scats1}, and SCOOT~\cite{scoot2}, have undergone widespread implementation in urban environments, significantly contributing to the mitigation of traffic congestion.

With the rapid advancement of artificial intelligence and the growing abundance of available traffic data, such as surveillance camera feeds in recent years, the pattern of TSC has undergone substantial evolution. Among these changes, reinforcement learning (RL) techniques have driven significant progress within TSC. For example, CoLight~\cite{colight} demonstrates exceptional performance and scalability in large-scale TSC, while AttendLight~\cite{attend}, another innovative model, exhibits versatility in handling various intersection topologies. These pioneering developments underscore the transformative potential of emerging technologies in reshaping the future landscape of traffic management at signalized intersections.

Generally, RL-based methodologies significantly enhance TSC performance through three primary approaches. First, some methods contribute to the field by ingeniously designing effective state representations or reward functions, as exemplified by PressLight~\cite{presslight} and Advanced-XLight~\cite{advanced}. These advancements aim to optimize decision-making processes within TSC, ensuring a more nuanced and responsive system. Second, developing advanced neural networks, as observed in FRAP~\cite{frap} and CoLight~\cite{colight}, significantly enhances transportation efficiency. These developments pave the way for more streamlined and adaptive control systems capable of responding dynamically to the complexities of urban traffic. 
Third, integrates advanced RL techniques, such as HiLight~\cite{hilight} and MetaLight~\cite{metalight}. These approaches explore new horizons in learning and adaptation, pushing the limits of optimizing traffic flow and alleviating congestion. 

Despite these remarkable advancements, prior studies still grapple with the challenge of insufficiently addressing inherent limitations within the existing TSC control framework. Contemporary advanced RL strategies for TSC predominantly employ a single-stage control framework. For each fixed action duration, an appropriate signal phase is determined, with the choice between maintaining the current phase or switching to a more suitable one. This mechanism mirrors control systems found in human-interactive games, such as Atari~\cite{atari}. However, such a single-stage control framework exhibits two primary limitations. First, the duration of each phase is significantly influenced by the fixed action duration~\cite{advanced,ql}, lacking sufficient flexibility and variability in phase durations. Second, the duration of each phase cannot be ascertained until another phase is actuated. Therefore, there exists a critical need to develop models capable of supporting dynamic phase durations. Recognizing and overcoming these challenges is essential for unlocking the full potential of RL-based methodologies in revolutionizing TSC.

This study introduces a novel two-stage framework named \textit{DynamicLight} to enhance the single-stage framework and achieve dynamic phase duration. The improvement involves integrating a duration control strategy that actively determines the phase duration, rather than allowing passive variation. Within such a new structure, one policy is dedicated to controlling the traffic phase, while another is responsible for determining the corresponding duration. This sophisticated two-stage approach promises to introduce a higher degree of adaptability and responsiveness to the dynamic nature of traffic conditions, marking a significant advancement in the realm of intelligent traffic management systems. The main contributions are organized as follows:
\begin{itemize}[leftmargin=*]
\item \textbf{Two-stage dynamic TSC framework}: 
Introducing DynamicLight, an efficient two-stage control framework employing a dual-policy mechanism. This framework seamlessly integrates phase selection and duration determination, allowing for dynamic phase durations in TSC.
\item \textbf{Robust scalability of DynamicLight}: 
Various DynamicLight variants are created by replacing the phase control strategy with an alternative one. These variants validate the effectiveness and robustness of our framework, highlighting its practical applicability.  
\item \textbf{Superior performance beyond state-of-the-art (SOTA) models}: Experimental results show that DynamicLight surpassed SOTA TSC models, establishing a new benchmark for advanced traffic control systems. 
\end{itemize}

\section{Related Work}
\label{sec:related}

\subsection{Traditional Methods}
\label{subsec:traditional}
In the realm of real-world TSC, commonly applied traditional methods exhibit a significant dependence on either manually crafted signal plans or rule-based systems.

FixedTime~\cite{fixedtime} is a traffic signal timing strategy that effectively regulates traffic signal operations by relying on predetermined values for cycle length, phase sequence, and phase split. GreenWave~\cite{greenwave} is designed to analyze applicable conditions of the Green-Wave traffic theory, employing a two-phase signal control concept for optimization. This strategy allows vehicles to pass through multiple intersections consecutively on green lights, optimizing traffic flow on main roads. Actuated control~\cite{sotl2013} introduced a self-organizing mechanism that dynamically responds to varying traffic conditions. This innovation has improved traffic flow by enabling traffic signals to autonomously adapt based on pre-defined rules and real-time traffic data. Adaptive control systems, such as SCATS ~\cite{scats1} and SCOOT~\cite{scoot2}, employed a decision-making process to select optimal traffic plans based on real-time data obtained from loop sensors. Widely embraced in large urban settings, such adaptive control systems significantly enhance traffic flow and responsiveness by dynamically adjusting to the prevailing traffic conditions. 

Recently, traditional optimization-based methodologies, such as Max Pressure~\cite{mp2013} and MaxQueueLength~\cite{ql}, employed max-pressure and max queue-length strategies to optimize TSC. These approaches have demonstrated significant efficacy in tackling complex congestion challenges at urban intersections, leading to a substantial enhancement in the overall efficiency of traffic management systems.

\subsection{RL-based Methods}
\label{subsec:rl}
Several RL-based methodologies enhanced TSC performance by designing effective state representations of reward functions. LIT~\cite{LIT} made significant strides in optimizing TSC by introducing a streamlined approach to state and reward design. This innovative methodology proved to be highly effective, surpassing the performance of IntelliLight~\cite{intellilight}. PressLight~\cite{presslight} advanced the capabilities of LIT and IntelliLight through the seamless integration of “pressure" into both the state and reward design. This integration significantly contributed to enhancing the overall TSC strategy, demonstrating its effectiveness in coordinating signals on arterial roadway networks. MPLight~\cite{mplight} enhanced FRAP~\cite{frap} by incorporating “pressure" in the state representation and reward function design.
AttentionLight~\cite{ql} employed queue length for both state representation and reward function, significantly surpassing FRAP. Advanced-XLight~\cite{advanced} introduced effective running vehicle number and traffic movement pressure as the state representations, demonstrating SOTA performance.

Furthermore, some RL-based methods have significantly enhanced TSC performance by developing sophisticated network structures. FRAP~\cite{frap} demonstrated exceptional skill in crafting phase features and adeptly capturing intricate relationships arising from phase competition in TSC. CoLight~\cite{colight} harnessed the capabilities of a graph attention network~\cite{gats}, specifically tailored to facilitate seamless cooperation at intersections, showcasing improved TSC efficacy. AttendLight~\cite{attend} utilized an attention network to adeptly manage diverse intersection topologies. 

Some other RL-based methodologies adopted advanced RL techniques to enhance model performance. DemoLight~\cite{demolight} utilized imitation learning~\cite{imitation} to accelerate learning. HiLight~\cite{hilight} enabled each strategy to learn a high-level policy, optimizing the objective locally using hierarchical RL~\cite{hierarchical}. MetaLight~\cite{metalight} utilized meta-learning~\cite{meta} to efficiently and robustly adapt to changing traffic scenarios.

All the aforementioned methods utilized a single-stage control framework. However, the duration of a phase is solely influenced by the action duration, leading to a lack of adequate variability. Moreover, the single-stage framework lacks the capability to pre-determine the duration of each stage before initiating the next one. In this study, we introduce DynamicLight, a two-stage framework designed to improve upon the single-stage framework and enable dynamic phase durations.

\section{Preliminaries}
\label{sec:preliminary}
\begin{figure}[ht]
\centering
\includegraphics[width=1\linewidth]{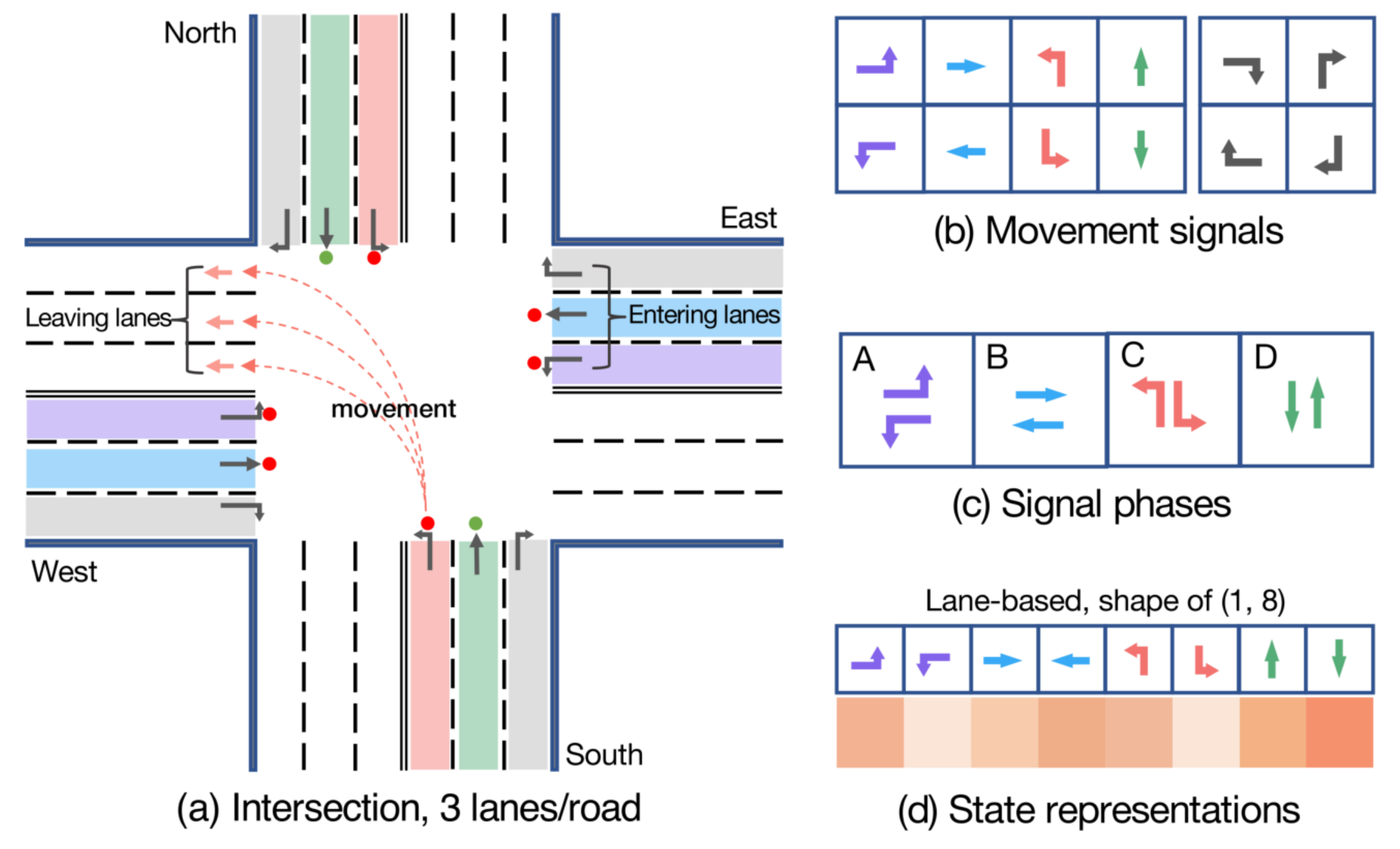}
\caption{Illustration of a standard intersection structure with four entry and four exit approaches (East, West, South, and North), each featuring three types of lanes (left, straight, and right). Subfigures depict (b) traffic movement signals, (c) signal phases, and (d) state representations for a comprehensive overview.}
\label{fig:intersection}
\end{figure}

\paragraph{Traffic network.} 
A typical representation of a traffic network involves a directed graph, with nodes corresponding to intersections and roads corresponding to edges. Figure~\ref{fig:intersection} (a) illustrates a standard intersection structure within this graph. Each road is composed of three types of lanes (i.e., turning left, going straight, and turning right), acting as the fundamental units facilitating vehicle movement and determining the trajectory of each vehicle passing through the intersection. An incoming lane serves as the entry point for vehicles approaching the intersection, orchestrating the initial flow of traffic. An outgoing lane provides a designated area for vehicles to seamlessly exit an intersection, thereby enhancing the overall efficiency of the traffic network. 

\paragraph{Traffic movements and phases.} 
A traffic movement refers to vehicles traveling at an intersection in a specific direction. In certain countries, vehicles making a right turn are allowed to proceed regardless of the signal but must come to a stop at a red light, as indicated by the black signals in Figure~\ref{fig:intersection} (b). Additionally, each intersection has its own phase settings. A signal phase comprises a set of permitted traffic movements. As illustrated in Figure \ref{fig:intersection} (c), each of the four signal phases controls two traffic movements that do not conflict with each other. Once a phase is activated, its duration is the period during which it remains active. To comprehensively reflect the traffic environment, the state representations are lane-based, as depicted in Figure~\ref{fig:intersection} (d).

\paragraph{Problem statement.} 
In a multi-intersection TSC system, each intersection is managed by an RL agent. An agent observes the environment and takes actions involving phase and duration, which lead to receiving a reward. The objective function for all agents is to learn an optimal policy that maximizes their cumulative rewards. For ease of deployment, certain agents are designed to handle various intersection topologies, ensuring adaptability to different configurations and enhancing the overall versatility of the implemented system.

\section{DynamicLight}
\label{sec:dynamiclight}
\begin{figure*}[t]
\centering
\includegraphics[width=1\textwidth]{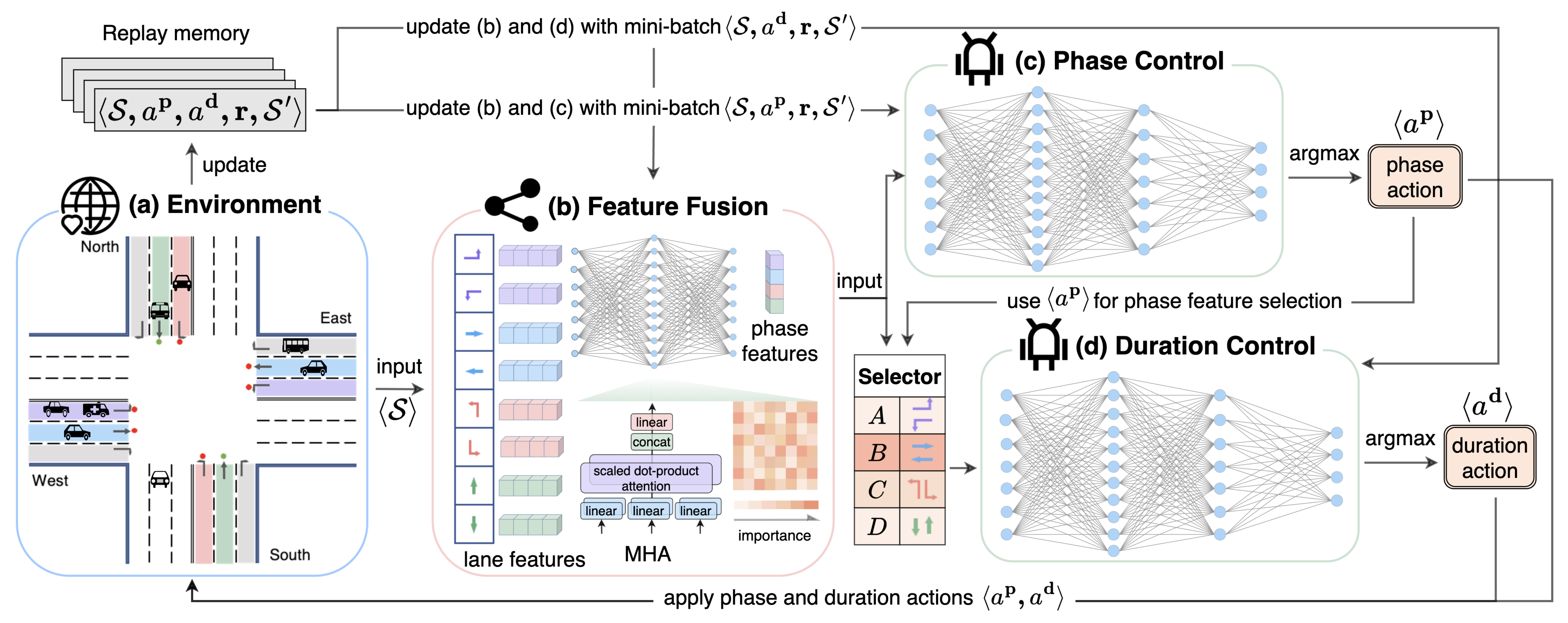}
\caption{Overview architecture of DynamicLight. \textbf{(a) The TSC environment} facilitates DynamicLight by providing state representations $\mathcal{S}$, executing received actions $\langle a^p, a^d\rangle$, and generating new states $\mathcal{S}^{\prime}$ and rewards $r$. It serves as the essential interface for interaction, enabling the seamless flow of information and feedback between the agent and its environment. These transition tuples $\langle\mathcal{S}, a^p, a^d, r, \mathcal{S}^{\prime} \rangle$ at an intersection are collected as the replay memory. \textbf{(b) Feature fusion} involves acquiring states from the environment and embedding them into lane features. Subsequently, the lane features undergo phase feature fusion through a multi-head self-attention (MHA) mechanism. \textbf{(c) Phase control} utilizes phase features as inputs and employs a deep network to approximate the Q-values. \textbf{(d) Duration control} selects the phase feature corresponding to the predicted phase action in (c) and embeds it to predict the Q-values. The phase action and duration action are determined using argmax operation. Note that the networks in (b) and (c) are updated with mini-batches $\langle\mathcal{S}, a^p, r, \mathcal{S}^{\prime}\rangle$ from the replay memory. Similarly, the networks in (b) and (d) are updated with mini-batches $\langle \mathcal{S}, a^d, r, \mathcal{S}^{\prime}\rangle$.}
\label{fig:model}
\end{figure*}

DynamicLight, a two-stage TSC framework, utilizes one deep Q-network for both the phase and duration control to dynamically adjust phase durations. Specifically, the phase control is responsible for determining the optimal traffic phase, while the duration control is tasked with determining the duration of the selected phase. Figure~\ref{fig:model} shows the overview architecture of DynamicLight. 

\subsection{DynamicLight Agent}
\label{subsec:control_agents}
\paragraph{State.}
Consider a TSC system with $N$ intersections $\mathcal{I}=$ $\{\mathcal{I}_1$, $\cdots$, $\mathcal{I}_N\}$. Here, $\mathcal{L}^{in}$ and $\mathcal{L}^{out}$ represent the sets of incoming and outgoing lanes, respectively, for a specific intersection. Seven state representations are utilized to describe the environment. Formally, let $\mathcal{S}_l=[s_{l,1}, \cdots, s_{l,7}]$ denote the set of all state descriptors, where $s_{l,i}$, $l\in\mathcal{L}^{in}$ represents the $i$-th state representation on lane $l$. These state representations include the current phase ($s_{l,1}$), queue length ($s_{l,2}$), effective running vehicle number ($s_{l,3}$), and the number of vehicles under the segmented road (four segments of 100 meters each, i.e., $s_{l,4}$ to $s_{l,7}$).

\paragraph{Action.} 
Define the phase and duration action spaces as $\mathcal{A}^p=\{a_1^p, a_2^p,\cdots, a_4^p\}$ and $\mathcal{A}^d =\{a_1^d,a_2^d,$ $\cdots,a_7^d\}$, respectively. Each element in $\mathcal{A}^p$ corresponds to a specific signal phase type (e.g., Type A, B, C, or D), while each element in $\mathcal{A}^d$ represents the duration time of a phase. In this study, we extensively explored the duration action space and ultimately selected $\mathcal{A}^d = \{10, 15, 20, 25, 30, 35, 40\}$ seconds in Appendix~\ref{apxsec:dspace}. At an intersection, the agent selects a phase action $a_i^p$ as its initial phase and subsequently maintains it for the duration of $a_j^d$. These two actions control the signal phase of the intersection, and the agent receives a reward based on its decisions. Through $N_t$ interactions, each agent learns and refines its control policies over time.

\paragraph{Reward.} 
Both the phase and duration controls utilize negative intersection queue length as their rewards, with the reward for controlling an intersection denoted as $r = -\sum s_{l,2}$. Intuitively, DynamicLight seeks to minimize the average travel time by maximizing the reward.

\subsection{Deep Q-Network Design}
\label{subsec:q_learning}
\paragraph{Feature fusion.}
The features of each state descriptor $s_{i,l}$ are initially embedded and concatenated to a lane feature:
\begin{align}
\mathbf{F}_l=\operatorname{Embed}( \operatorname{Embed}(s_{l,1}) \oplus \cdots \oplus \operatorname{Embed}(s_{l,7})),
\end{align}
where $\oplus$ denotes the concatenation operation.  Various feature fusion methods, including addition~\cite{frap}, embedding with a multi-layer perceptron (MLP), and multi-head self-attention (MHA)~\cite{attention}, were explored in Appendix ~\ref{apxsec:fusion}. Finally, MHA was chosen due to its superior performance. Since each phase comprises two lanes ($\mathbf{F}_{l_1}$ and $\mathbf{F}_{l_2}$ as illustrated in Figure~\ref{fig:model} (c)), the averaged feature fusion for phase $p$ can be calculated by
\begin{align}
\mathbf{F}^p = \operatorname{Mean}\left(\operatorname{MHA}(\mathbf{F}_{l_1} \oplus \mathbf{F}_{l_2})\right).
\end{align}
Note that the fused phase feature $\mathbf{F}^p$ serves as the input for Q-value prediction in both phase and duration controls.

\paragraph{Q-value prediction.}
All the fused phase features are modeled with MHA to capture their correlations, and the correlated features are embedded to generate Q-values for the phase control. Subsequently, the phase action with the maximum Q-value is selected. In practice, the phase control needs to complete its task before the duration control, as selecting an appropriate duration depends on the determined phase.

Next, the fused phase feature $\mathbf{F}^p$ and the pre-determined phase action serve as inputs to the duration control. The four features are concatenated, and the result is multiplied by the representation of the phase action to extract the corresponding phase feature. The extracted phase feature is further embedded to obtain Q-values. Finally, the duration action with the maximum Q-value is selected. The final embedding is shared across different phases, enabling it to handle various types of phase features and benefit from shared experiences (see details in Appendix~\ref{apxsec:share}). For more information on DynamicLight architecture, see Appendix~\ref{apxsec:dynnet}.

\subsection{Training Procedure}
\label{subsec:training}
For the TSC system with $N$ intersections, the control process for the $n$-th intersection can be represented by a Markov decision process, denoted by the tuple $\mathcal{M}_n=\langle \mathcal{S}, a^p, a^d, r, \gamma\rangle$, where $\mathcal{S}$ denotes a set of state representations, $a^p\in \mathcal{A}^{p}$, $a^d\in \mathcal{A}^{d}$, $r$ is a reward function, and $\gamma$ represents a discount factor at the $n$-th intersection. Then, the collected replay memory can be denoted as $\mathcal{M} = \bigcup_{n\in N}\mathcal{M}_n$. 

Let $\bm{\phi}$ represent the parameters of the feature fusion network, $\bm{\theta}_1$ and $\bm{\theta}_2$ denote the parameters of neural networks for action and duration controls, respectively. These parameters are utilized to approximate the Q-value function of the RL agent. Then, the temporal difference loss for phase control can be calculated as follows:
\begin{small}
\begin{align}  
\mathcal{L}(\bm{\phi},\bm{\theta}_1)&=\mathbb{E}_{(\mathcal{S}, a^p, r,  \mathcal{S}^{\prime}, \gamma)\sim \mathcal{M}}\left[\left(y-Q(\mathcal{S}, a^p; \bm{\phi}, \bm{\theta}_1)\right)^2\right],
\end{align}
\end{small}
where $Q(\cdot)$ represents the Q-value function, and $y$ is the target Q-value, which is calculated as
\begin{align} 
y = r + \gamma \max _{a^{p\prime}} Q(\mathcal{S}^{\prime}, a^{p\prime}; \bm{\phi}^-,\bm{\theta}_1^-).
\end{align}
Here, $\mathcal{S}^{\prime}$ and $a^{p\prime}$ denote the next state and action of phase control, respectively, and $\bm{\phi}^-$ and $\bm{\theta}_1^-$ represent the parameters of their target networks~\cite{double}. Finally, the gradients of phase control can be updated by
\begin{align}  
\bm{\phi} &\leftarrow \bm{\phi} - \alpha \nabla_{\bm{\phi}} \mathcal{L}(\bm{\phi},\bm{\theta}_1),
\end{align}
and
\begin{align}
\bm{\theta}_1 &\leftarrow \bm{\theta}_1 - \beta \nabla_{\bm{\theta}_1} \mathcal{L}(\bm{\phi},\bm{\theta}_1),
\end{align}
where $\alpha$ and $\beta$ represent the learning rates for the feature fusion network and phase control strategy, respectively. Similarly, the parameters of DynamicLight's duration control strategy $\bm{\theta}_2$ can be updated in the same manner.

Algorithm \ref{alg:dyn} illustrates the training procedure of DynamicLight. To enhance the training of DynamicLight, a variety of techniques have been employed. The impact of these techniques on the training effectiveness is detailed in Appendix~\ref{apxsec:training}. Furthermore, during the training of phase control, a fixed duration action was utilized, which also influenced the model's performance. For further information on this aspect, refer to Appendix~\ref{apxsec:fixd}.

\begin{algorithm}[t]
\caption{Training procedure of DynamicLight.}
\label{alg:dyn}
\textbf{Initialization}: the Parameters of $\bm{\phi}$, $\bm{\theta}_1$ and $\bm{\theta}_2$, discount factor $\gamma$, replay buffer $\mathcal{M}$, and the time steps $T_1$ and $T_2$ for DynamicLight updating
\begin{algorithmic}[1]
\FOR{ $t=1 \rightarrow T$}
\STATE Select phase and duration actions according to their policies and receive reward $r$, new state $\mathcal{S}^{\prime}$
\STATE Store the transition tuple $\langle\mathcal{S}, a^p, a^d, r, \mathcal{S}^{\prime}\rangle$ to $\mathcal{M}$
\STATE Sample a random mini-batch from $\mathcal{M}$
\IF{$t<T_1$} 
\STATE Update parameters $\bm{\phi}$ and $\bm{\theta}_1$ using Eqs (2)-(4)
\ELSIF{$t<T_2$}
\STATE Fix parameter $\bm{\phi}$ and update parameters $\bm{\theta}_2$
\ELSE
\STATE Update parameters $\bm{\phi}$, $\bm{\theta}_1$, and $\bm{\theta}_2$
\ENDIF
\STATE Update the parameters of target networks
\STATE $t\leftarrow t+1$
\STATE $\mathcal{S} \leftarrow \mathcal{S}^{\prime}$
\ENDFOR

\end{algorithmic}
\end{algorithm}

\section{Experiments}
\label{sec:exp}

CityFlow~\cite{cityflow}, an open-source platform supporting large-scale TSC simulations, utilizes a structure where each green signal is succeeded by a five-second red signal to facilitate signal phase transition. In this section, we primarily conducted extensive experiments on CityFlow to answer the following research questions (RQ):
\begin{itemize}[leftmargin=*]
\item \textbf{RQ1}: How does the performance of DynamicLight compare with that of SOTA approaches? 
\item \textbf{RQ2}: How does the scalability of DynamicLight? 
\item \textbf{RQ3}: Is the duration control strategy effective?
\item \textbf{RQ4}: Can DynamicLight be effectively applied to various intersection topologies?
\end{itemize}

\begin{figure}[t]
\centering
\includegraphics[width=1\linewidth]{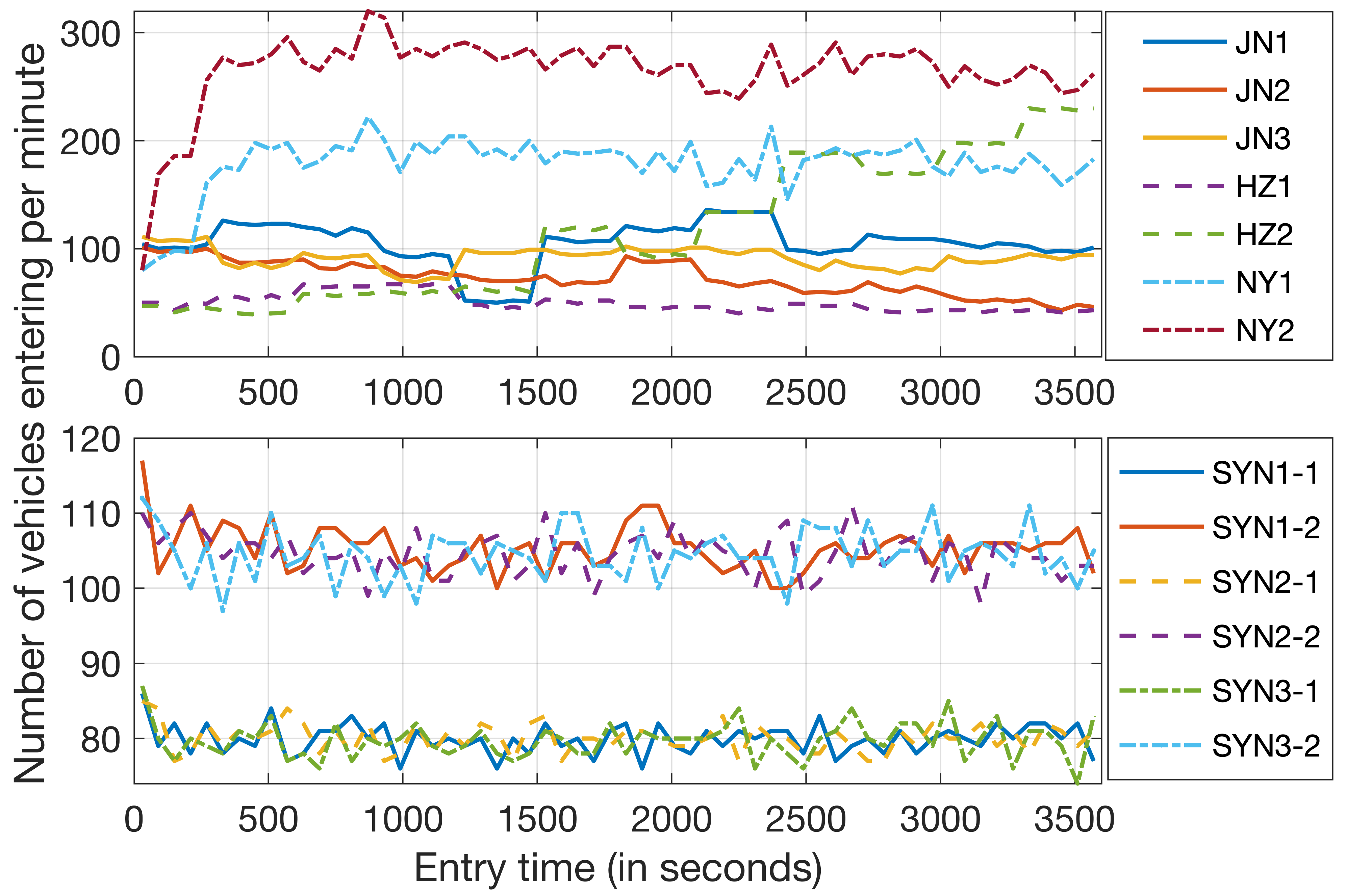}
\caption{Visual depiction of the number of vehicle entries in both real-world and synthetic datasets.}
\label{fig:data}
\end{figure}

\paragraph{Datasets.} Both real-world and synthetic datasets were utilized to validate the effectiveness of the DynamicLight framework in our experiments. Each traffic dataset includes a road network dataset, providing details about the traffic network, and a traffic flow dataset, illustrating the movement of vehicles across the network along predetermined routes.
\begin{itemize}[leftmargin=*]
\item \textbf{Real-world datasets} consist of seven traffic flow datasets. Three datasets from Jinan (JN) are labeled as JN1, JN2, and JN3. Two datasets from Hangzhou (HZ) are denoted as HZ1 and HZ2. Additionally, two datasets from New York (NY) are identified as NY1 and NY2. The intersections in the road networks of JN, HZ, and NY share the same topologies as depicted in Figure~\ref{fig:intersection} (a).

\item \textbf{Synthetic datasets} comprise six traffic flow datasets for three distinct intersection topologies, denoted as SYN1-1, SYN1-2, SYN2-1, SYN2-2, SYN3-1, and SYN3-2. Their road network is identical to that in JN. Each traffic flow dataset was randomly generated based on statistical analysis of real-world datasets with turn ratios of 10\% (left), 60\% (straight), and 30\% (right). Figure \ref{fig:case} (a) shows the three intersection topologies. 
\end{itemize} 
Figure~\ref{fig:data} visually presents the distinct arrival rates and travel patterns exhibited by each traffic flow dataset, encompassing both real-world and synthetic datasets. The diversity observed across these datasets highlights the robustness and comprehensiveness of our experiments.

\paragraph{Hyperparameter settings.}
The Adam optimizer was employed as an optimization function for both phase and duration control with a learning rate of $1\mathrm{e}{-3}$. The memory size of the replay buffer and the discount factor $\gamma$ were set to 12,000 and 0.8, respectively. DynamicLight underwent training for 100 epochs per episode, with a total of 200 episodes. In both the training and testing phases, each episode simulated traffic conditions for 60 minutes.

\paragraph{Metrics.}
Average travel time (ATT), a commonly used performance measure, such as FRAP~\cite{frap} and Advanced-XLight~\cite{advanced}, was utilized as the evaluation metric in this study. To enhance the reliability of our experimental assessment, we averaged ATT over the final ten testing episodes, conducted three sets of independent experiments, and reported the average results.

\subsection{Evaluation Results (RQ1)}
\label{subsec:exp_res}
\begin{table*}[t]
\setlength\tabcolsep{5.4pt}
\centering
\begin{threeparttable}
\caption{Evaluation results of ATT metric (in seconds) for DynamicLight vs. other baseline models.}
\label{tab:overall}
\begin{tabular}{lccc|cc|cc}\toprule
\multirow{2}{*}[-0.4em]{Model}&\multicolumn{3}{c}{JN dataset}&\multicolumn{2}{c}{HZ dataset}&\multicolumn{2}{c}{NY dataset}\\\cmidrule {2-8} 
&JN1&JN2&JN3&HZ1&HZ2&NY1&NY2\\\midrule
FixedTime~\cite{fixedtime}&429.27&370.34&384.89&497.87&408.31&1507.12&1733.30\\
Max Pressure~\cite{mp2013}&274.99&246.41&244.63&289.55&349.85&1179.55&1536.17\\
Max QueueLength~\cite{ql}&268.87&240.02&238.51&284.32&325.44&1197.59&1551.46\\\midrule
FRAP~\cite{frap}&299.56&268.57&269.20&308.73&355.80&1192.23&1470.51\\
MPLight~\cite{mplight}&297.68&274.32&268.00&313.16&355.35&1321.40&1642.05\\
PRGLight~\cite{prglight}&291.27&257.52&261.74&301.06&369.98&1283.37&1472.73\\
CoLight~\cite{colight}&271.17&251.22&248.87&300.07&339.76&1065.64&1367.54\\
AttentionLight~\cite{ql}&254.82&239.68&236.62&283.64&316.38&1013.78&1401.32\\
Advanced-CoLight~\cite{advanced}&246.42&233.72&229.47&271.64&313.51&970.05&\textbf{1300.62}\\\midrule
\textbf{DynamicLight}&\textbf{235.95}&\textbf{221.29}&\textbf{218.66}&\textbf{262.24}&\textbf{303.23}&\textbf{956.59}&1334.92\\
Improvement&$\uparrow$ 4.25\%&$\uparrow$ 5.32\%&$\uparrow$ 4.71\%&$\uparrow$ 3.46\%&$\uparrow$ 3.28\%&$\uparrow$ 1.39\%&$\downarrow$ 2.57\%\\\bottomrule
\end{tabular}
\begin{tablenotes}
\footnotesize
\item[$\star$] The values highlighted in bold represent the maximum values among all compared models.
\end{tablenotes}
\end{threeparttable}
\end{table*}
Various TSC optimization methods were employed to facilitate a comprehensive evaluation. FixedTime, Max Pressure, and Max QueueLength are representative of traditional TSC methods. In contrast, FRAP, MPLight, PRGLight, CoLight, AttentionLight, and Advanced-CoLight represent superior RL-based TSC optimization methods. Note that Advanced-CoLight was the SOTA model, achieving the best performance among all the baseline models until now. 

Table~\ref{tab:overall} presents the evaluation results of the ATT metric for DynamicLight and the aforementioned baseline models across real-world datasets. Our proposed DynamicLight surpasses Advanced-CoLight, establishing a new SOTA performance for TSC. In particular, In Table~\ref{tab:overall}, DynamicLight outperformed Advanced-CoLight in ATT results on the JN and HZ datasets, exhibiting improvements of 4.25\% on JN1, 5.32\% on JN2, 4.71\% on JN3, and improvements of 3.46\% and 3.28\% on HZ1 and HZ2, respectively. For the results on the NY dataset, DynamicLight demonstrated a 1.39\% improvement on NY1 and a reduction of 2.57\% on NY2. Overall, DynamicLight outperforms all baselines on six out of seven real-world datasets, highlighting the effectiveness of the proposed two-stage framework and setting a new benchmark for TSC optimization.

\subsection{Scalability Exploration (RQ2)}
\label{subsec:scalability}
To address RQ2, we integrated the proposed two-stage framework with various single-stage models to explore the scalability of DynamicLight. Refer to Appendices \ref{apxsec:variants} and \ref{apxsec:lite} for additional details of DynamicLight variants.
\begin{itemize}[leftmargin=*]
\item \textbf{DynamicLight-Rand} applies a random policy (Rand) as the phase control policy, with the negative queue length as the reward.
\item \textbf{DynamicLight-FT} uses FixedTime (FT) as the phase control strategy and the negative queue length as the reward.
\item \textbf{DynamicLight-MP} employs Max Pressure (MP) as the phase control policy, utilizing the negative absolute pressure as the reward function in the duration control strategy to maintain a consistent optimization target.
\item \textbf{DynamicLight-MQL} adopts max queue-length (MQL) for the phase control and employs the negative queue length as the reward.
\item \textbf{DynamicLight-Lite} originates from DynamicLight-MQL, streamlining the network of the duration control and reducing the total parameters from $6,376$ to 19.
\end{itemize}
\begin{figure}[t]
\centering
\includegraphics[width=1\linewidth]{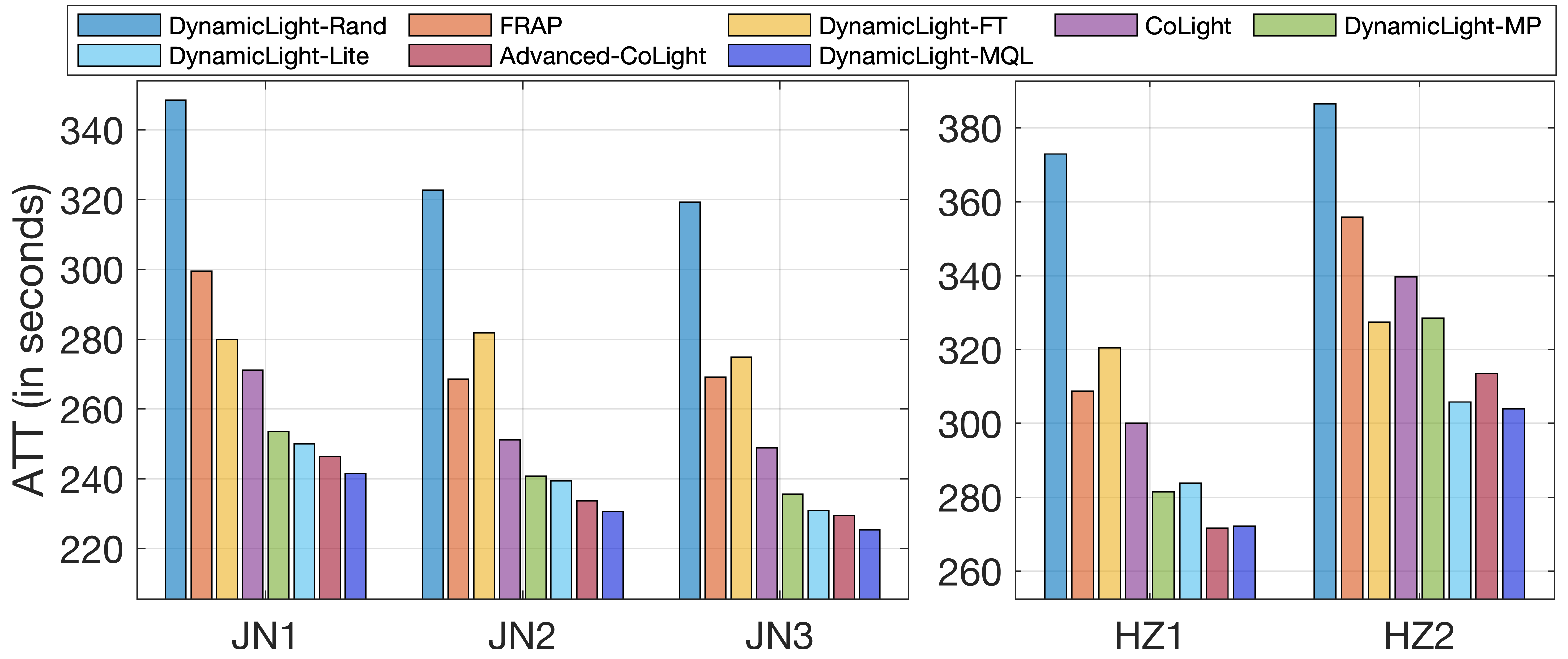}
\caption{Performance comparison of DynamicLight variants (ATT in seconds).}
\label{fig:scal}
\end{figure}
The results presented in Figure~\ref{fig:scal} illustrate the performance of DynamicLight variants. DynamicLight-MQL outperforms the previous SOTA model, Advanced-CoLight. The superiority of DynamicLight-FT over FRAP, despite having a fixed cyclical phase order, emphasizes the crucial role of the dynamic two-stage framework. DynamicLight-Lite, representing a trade-off between performance and computational cost, achieves comparable results to Advanced-CoLight with only 19 parameters, further affirming the effectiveness of the two-stage framework. The learning prowess and efficiency of the duration control strategy are underscored by DynamicLight-Rand, which exhibits satisfactory performance with random selection as the phase control strategy. Overall, these results validate the robust scalability of DynamicLight.

\subsection{Ablation Studies (RQ3)}
\label{subsec:ablation}
\paragraph{Effects of duration control strategy.}
\begin{table}[t]
\renewcommand{\arraystretch}{-0.3}
\setlength\tabcolsep{3pt}
\caption{Performance comparisons of DynamicLight and its variants with and without duration control (ATT in seconds).}
\label{tab:ablation}
\centering
\begin{tabular}{lcc|c}\toprule
\multirow{2}{*}[0.4em]{Model}&\multicolumn{2}{c}{JN dataset}&\multicolumn{1}{c}{HZ dataset}\\\cmidrule{2-4}&JN1&JN2&HZ1\\\midrule

w/o duration&244.28&231.82&270.29\\
\multirow{2}{*}[0.8em]{DynamicLight}&\makecell{\textbf{235.95}\\\footnotesize($\uparrow$ 3.41\%)}&\makecell{\textbf{221.29}\\\footnotesize($\uparrow$ 4.54\%)}&\makecell{\textbf{262.24}\\\footnotesize($\uparrow$ 2.98\%)}\\\midrule

w/o duration&559.49&506.72&560.08\\
\multirow{2}{*}[0.8em]{DynamicLight-Rand}&\makecell{\textbf{348.48}\\\footnotesize($\uparrow$ 37.71\%)}&\makecell{\textbf{322.78}\\\footnotesize($\uparrow$ 36.03\%)}&\makecell{\textbf{372.91}\\\footnotesize($\uparrow$ 33.42\%)}\\\midrule

w/o duration&429.27&370.34&497.87\\
\multirow{2}{*}[0.8em]{DynamicLight-FT}&\makecell{\textbf{279.99}\\\footnotesize($\uparrow$ 34.78\%)}&\makecell{\textbf{281.87}\\\footnotesize($\uparrow$ 23.89\%)}&\makecell{\textbf{320.49}\\\footnotesize($\uparrow$ 35.63\%)}\\\midrule

w/o duration&274.99&246.41&289.55\\
\multirow{2}{*}[0.8em]{DynamicLight-MP}&\makecell{\textbf{253.52}\\\footnotesize($\uparrow$ 7.81\%)}&\makecell{\textbf{240.77}\\\footnotesize($\uparrow$ 2.29\%)}&\makecell{\textbf{281.45}\\\footnotesize($\uparrow$ 2.80\%)}\\\midrule

w/o duration&268.87&240.02&284.32\\
\multirow{2}{*}[0.8em]{DynamicLight-MQL}&\makecell{\textbf{241.56}\\\footnotesize($\uparrow$ 10.16\%)}&\makecell{\textbf{230.63}\\\footnotesize($\uparrow$ 3.91\%)}&\makecell{\textbf{272.19}\\\footnotesize($\uparrow$ 4.27\%)}\\\bottomrule
\end{tabular}
\end{table}
To evaluate the effectiveness of the duration control strategy, we conducted performance comparisons for DynamicLight and its variants using real-world datasets. In the scenario without the duration control (i.e., w/o duration), the TSC is considered a single-stage framework with only the phase control, where the action duration time is fixed at 15 seconds. Table~\ref{tab:ablation} and Table~\ref{tab:apx_ablation} show the impacts of their duration controls. Compared to DynamicLight without a duration control, DynamicLight with a two-stage framework improved ATT by 3.41\%, 4.54\%, and 2.98\% on the JN1, JN2, and HZ1 datasets. Similarly, the ATT of the four variants of DynamicLight (DynamicLight-MQL, DynamicLight-MP, DynamicLight-FT, and DynamicLight-Rand) using the two-stage framework all exceed those of the single-stage framework without duration control. In particular, for DynamicLight-Rand, the ATT demonstrated an improvement of more than 33\% across the three real-world datasets. Overall, DynamicLight and its variants exhibit superior performance, affirming the efficacy of the proposed two-stage framework.

\paragraph{Learning capabilities of duration control strategy.} To further investigate the effects of the duration control strategy, FixedTime was set as the phase control strategy for DynamicLight and its variants after training, to eliminate the impact of the phase strategy on performance in testing. Table~\ref{tab:cap} and Table~\ref{tab:apx_cap} present the evaluation results of their learning capabilities on real-world datasets. As shown in Table~\ref{tab:cap}, the ATT of the duration control was significantly influenced by the selection of phase controls during the training phase. DynamicLight and DynamicLight-Lite achieved the two minimum ATT values on the JN1 and JN2 datasets, specifically 263.90 and 245.24, respectively, and DynamicLight-MQL obtained the minimum ATT (309.14) on the HZ1 dataset. These results illustrated the effectiveness of the duration control, demonstrating its ability to outperform RL-based methods even without optimizing phase order.
\begin{table}[t]
\caption{Examining the learning capability of the duration control strategy for DynamicLight and its variants with maintaining the duration strategy in training and changing phase strategy to FixedTime in testing (ATT in seconds).}
\label{tab:cap}
\centering
\begin{tabular}{lcc|c}\toprule
\multirow{2}{*}[-0.4em]{Model}&\multicolumn{2}{c}{JN dataset}&\multicolumn{1}{c}{HZ dataset}\\\cmidrule{2-4}
&JN1&JN2&HZ1\\\midrule
DynamicLight&\textbf{263.90}&\textbf{245.24}&328.31\\
DynamicLight-Rand&301.82&277.65&322.79\\
DynamicLight-FT&279.99&281.87&320.49\\
DynamicLight-MP&283.80&264.45&310.69\\
DynamicLight-MQL&270.32&250.09&\textbf{309.14}\\
DynamicLight-Lite&277.61&254.02&506.53\\\bottomrule
\end{tabular}
\end{table}

\subsection{Case Studies on Intersection Topologies (RQ4)}
\label{subsec:case_study}
\begin{figure}[ht]
\centering
\includegraphics[width=1\linewidth]{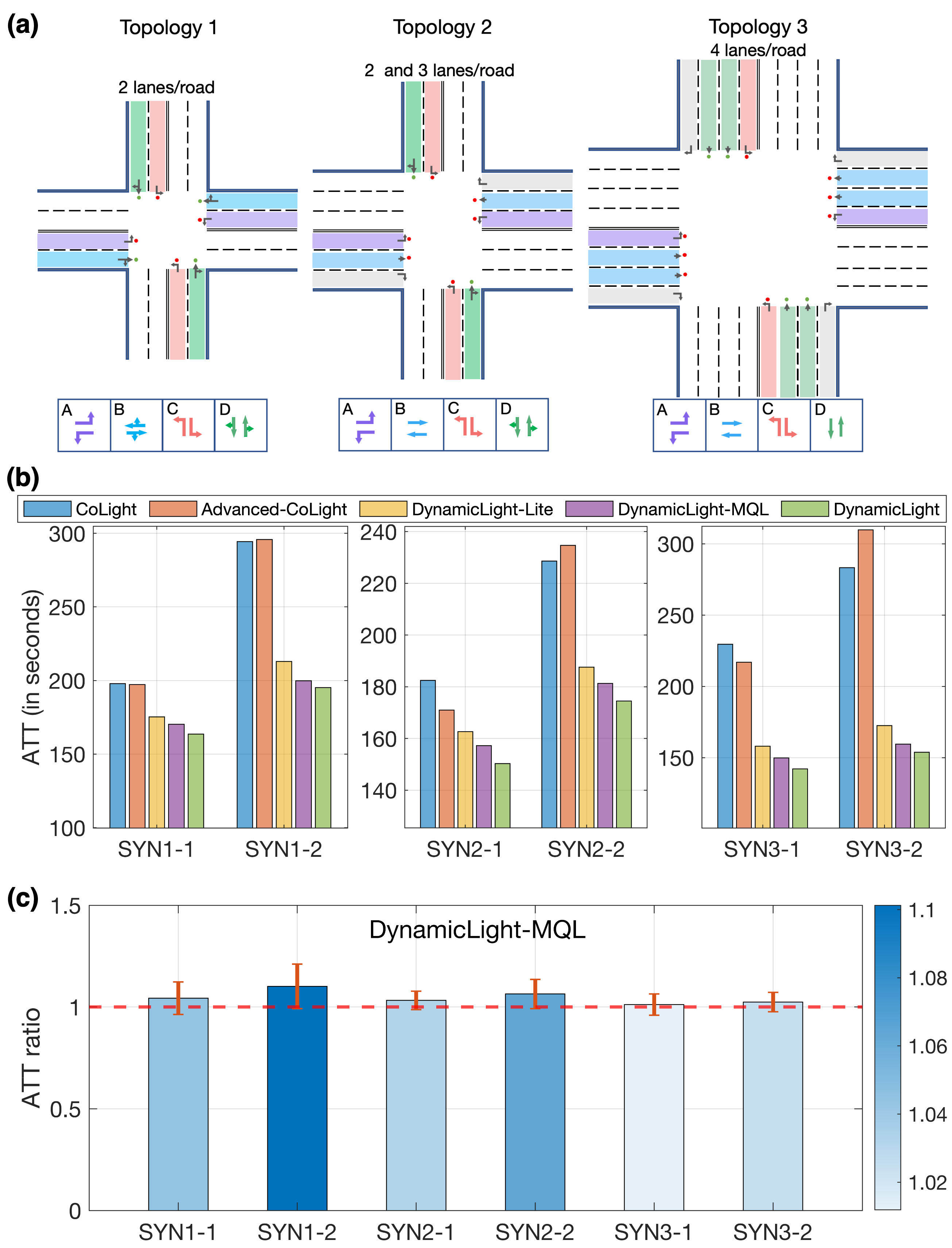}
\caption{(a) Illustration of three different intersection topologies. (b) Performance comparison of DynamicLight variants with baseline models. (c) Comparison of ATT ratio on synthetic datasets, the error bars represent a 95\% confidence interval for ATT ratio.}
\label{fig:case}
\end{figure}
To assess the capabilities of DynamicLight and its variants in handling various traffic scenarios, three sets of synthetic datasets were utilized, each featuring distinct intersection topologies as illustrated in Figure~\ref{fig:case} (a). 

Figure~\ref{fig:case} (b) and Table~\ref{tab:apx_syn} compare the performance of DynamicLight variants with CoLight and Advanced-CoLight. The results indicate that DynamicLight, DynamicLight-MQL, and DynamicLight-Lite exhibit satisfactory performance, with their ATTs surpassing all baseline methods, and ranking in the top three among all models.  Note that TSC models, such as FRAP and MPLight, lack the capabilities to effectively handle such intersections. In contrast, DynamicLight and its variants achieved SOTA performance on the three synthetic datasets, highlighting their exceptional capabilities in handling various intersection topologies. 

Figure~\ref{fig:case} (c) illustrates the ATT ratio of DynamicLight-MQL on the synthetic datasets. DynamicLight-MQL demonstrates effective adaptation to various intersection topologies, indicating its potential for training on one topology (Topology 1 in Figure~\ref{fig:case} (a)) and successfully transferring that knowledge to others (Topologies 2 and 3). This transferability allows for training the model on a small scale in the real world to reduce costs while deploying it broadly.

\section{Conclusion}
\label{sec:conc}
This study introduced a two-stage TSC framework called DynamicLight, utilizing one deep Q-network to dynamically adjust phase durations. DynamicLight begins with a phase control responsible for determining the optimal traffic phase and subsequently employs a duration control to determine the corresponding phase duration. Extensive experiments were conducted on both real-world and synthetic datasets to validate the effectiveness of DynamicLight. The experimental results demonstrate that DynamicLight achieved SOTA performance compared to all baselines, setting a new benchmark for advanced traffic control systems. Furthermore, DynamicLight variants were employed to verify the robustness of our framework, highlighting its significant potential for real-world deployment.

DynamicLight has two main limitations. Firstly, its training process, involving the cooperation of two distinct control strategies, demands extensive calculations to achieve optimal results. Secondly, DynamicLight lacks the capability to incorporate information from neighboring intersections, thus missing out on the benefits of coordinated intersection cooperation. In the future, we plan to develop an enhanced version of DynamicLight that can exchange information with neighboring intersections to improve performance.

\section*{Impact Statements}
In the realm of RL-based TSC methodologies, the two-stage DynamicLight framework and its variants exhibit superior suitability for real-world deployment compared to conventional one-stage control approaches. This assertion is underpinned by three pivotal aspects:
\begin{itemize}[leftmargin=*]
\item Definitive Phase Duration: Unlike traditional one-stage control systems, DynamicLight provides the capability to prescribe fixed phase durations. This feature enhances predictability and consistency in traffic management, presenting a crucial factor in real-world traffic scenarios.
\item Optimized Cyclical Control: The two-stage architecture of DynamicLight facilitates optimized cyclical control, a feature not supported by traditional one-stage RL-based methods. This optimization is integral for deploying control systems in real-world settings that are readily acceptable and user-friendly.
\item Versatility Across Diverse Intersection Topologies: A significant advantage of DynamicLight's variants is their adaptability to various intersection topologies. This versatility is paramount in real-world settings where intersection configurations are not uniform and require a flexible control system.
\end{itemize}
Beyond focusing solely on model performance metrics, DynamicLight and its variants prioritize practical applicability. DynamicLight aims to accelerate the integration and advancement of RL methodologies in real-world traffic control applications, bridging the gap between theoretical models and their practical deployment in dynamic and unpredictable urban traffic environments.

\bibliography{reference}
\bibliographystyle{icml2024}

\clearpage
\onecolumn
\setcounter{secnumdepth}{2}
\numberwithin{figure}{section}
\numberwithin{table}{section}
\numberwithin{equation}{section}
\begin{appendices}


\section{Model Study Details}
In our model study, we employ both DynamicLight and DynamicLight-MQL (see Appendix~\ref{apxsec:variants}) to gain comprehensive insights. While DynamicLight is explored to understand its training dynamics, DynamicLight-MQL is adopted as a foundational model due to its simplicity, efficiency, and notably greater stability compared to traditional RL models. This choice effectively reduces the impact of instability factors often encountered in RL-based approaches.
\subsection{Duration Action Space Survey}
\label{apxsec:dspace}
To thoroughly assess the flexibility of duration control, we evaluate DynamicLight-MQL across five distinct sets of duration action spaces, described as follows: $\mathcal{D}_1=\{10, 20, 30, 40\}$ seconds; $\mathcal{D}_2=\{10, 15, 20, 25, 30, 35, 40\}$ seconds; $\mathcal{D}_3=\{10, 13, 16, 19, 22, 25, 28, 31, 34, 37, 40\}$ seconds; $\mathcal{D}_4=\{10, 15, 20, 25, 30, 35, 40, 45, 50, 55, 60\}$ seconds;  and $\mathcal{D}_5 =\{10, 15, 20\}$ seconds.
The performance of DynamicLight-QML under these varying duration action spaces is detailed in Table~\ref{apx:dspace}. For our study, $\mathcal{D}_2$ is chosen as the default duration action space. This selection is based on its balanced range and performance, offering a comprehensive view of the model's capabilities in handling diverse traffic scenarios.

\begin{table}[H]
\caption{Performance comparison of DynamicLight-MQL across different duration action spaces (ATT in seconds).}
\label{apx:dspace}
    \centering 
    \begin{tabular}{cccc|cc}
    \toprule
    \multirow{2}{*}[-0.5em]{Duration action space}&\multicolumn{3}{c}{JN dataset}&\multicolumn{2}{c}{HZ dataset}\\\cmidrule{2-6}&JN1&JN2&JN3&HZ1&HZ2\\\midrule
    $\mathcal{D}_1$ &$ 245.01$ & $ 230.77$& $ 228.00$&$272.71 $ &$303.96 $\\
    $\mathcal{D}_2$ &$ 241.56$ & $230.63 $& $\mathbf{225.31} $&$ 272.19$ &$\mathbf{303.92} $\\
    $\mathcal{D}_3$ &$ \mathbf{241.47}$ & $229.38 $&$227.88 $& $273.64 $ &$ 305.14$\\
    $\mathcal{D}_4$ &$ 242.73$ & $ 229.53$&$226.14 $& $\mathbf{271.91} $ &$ 304.61$\\
    $\mathcal{D}_5$ &$ 247.43$ & $ \mathbf{228.83}$& $226.14 $&$272.37 $ &$311.63 $\\
    \bottomrule
    \end{tabular}
\end{table}

\subsection{Feature Fusion Methods Selection}
\label{apxsec:fusion}
We conducted evaluations on four distinct phase feature fusion methods, each with its unique approach to integrating lane features for phase representation:
\begin{itemize}[leftmargin=*]
    \item \textbf{Fusion1:} This method employs multi-head self-attention (MHA)~\cite{attention} to process features from participating lanes, followed by averaging these features to derive the phase feature. DynamicLight and its variants, excluding DynamicLight-Lite, utilize this method.
    \item \textbf{Fusion2:} In this approach, the features of participating lanes are directly added together to form the phase feature. This method is adopted by FRAP~\cite{frap} and DynamicLight-Lite.
    \item \textbf{Fusion3:} This method applies MHA to both the participating lane features and their average to obtain the phase feature. AttendLight~\cite{attend} uses this fusion technique.
    \item \textbf{Fusion4:} Here, the participating lane features are first concatenated along the final axis, and then a single MLP is used to embed them into a uniform dimension.
\end{itemize}
The performance of DynamicLight-MQL with these different phase feature fusion methods is detailed in Table~\ref{apx:fusion}. Our findings indicate that DynamicLight-QML achieves optimal performance with Fusion1. Consequently, we have selected Fusion1 as the default method in our neural network design, considering its effectiveness in integrating lane features for phase representation.

\begin{table}[H]
\caption{Performance comparison of DynamicLight-MQL across different phase feature fusion methods (ATT in seconds).}
\label{apx:fusion}
    \centering 
    \begin{tabular}{cccc|cc}
    \toprule
    \multirow{2}{*}[-0.5em]{Feature fusion method}&\multicolumn{3}{c}{JN dataset}&\multicolumn{2}{c}{HZ dataset}\\\cmidrule{2-6}&JN1&JN2&JN3&HZ1&HZ2\\\midrule
    Fusion1 & $\mathbf{241.56}$ &$\mathbf{230.63}$ & $\mathbf{225.31}$& $\mathbf{272.19}$&$\mathbf{303.92}$ \\
    Fusion2 & $244.07$&$235.01$ &$231.57$ & $276.92$&$309.91$ \\
    Fusion3 & $242.09$& $228.42$&$225.69$ &$274.16$ &$306.36$ \\
    Fusion4 & $245.39$& $233.66$& $228.17$&$276.87$ &$306.97$ \\
    \bottomrule
    \end{tabular}
\end{table}

\subsection{Effects of Parameter Sharing for Duration Q-value Prediction}
\label{apxsec:share}
The component responsible for predicting duration Q-values in our model is designed to be potentially shared across different phases. To understand the impact of this design choice, we conducted evaluations both with and without parameter sharing in this module. The results of these evaluations are presented in Table~\ref{apx:share}. Our findings reveal that sharing the parameters of the duration Q-value prediction module can indeed enhance the overall performance of the model. This suggests that parameter sharing in this context not only streamlines the model but also contributes positively to its effectiveness.

\begin{table}[H]
\caption{Performance comparison of DynamicLight-MQL with and without parameter sharing (ATT in seconds).}
\label{apx:share}
    \centering 
    \begin{tabular}{lccc|cc}
    \toprule
    \multirow{2}{*}[-0.5em]{Config}&\multicolumn{3}{c}{JN dataset}&\multicolumn{2}{c}{HZ dataset}\\\cmidrule{2-6}&JN1&JN2&JN3&HZ1&HZ2\\\midrule
    w/o sharing &$245.30$&$ 233.38$ &$227.73$ &$275.05$ & $310.38$ \\
    w sharing & $\mathbf{241.56}$ &$\mathbf{230.63}$ & $\mathbf{225.31}$& $\mathbf{272.19}$&$\mathbf{303.92}$ \\
    \bottomrule
    \end{tabular}
\end{table}
\subsection{DynamicLight Network Details}
\label{apxsec:dynnet}
The architecture of the DynamicLight network is outlined as follows: 
\begin{itemize}[leftmargin=*]
    \item \textbf{Feature fusion.} Each incoming lane's characteristics, denoted as $s_{l,i}$ (where $ i \in \mathbb{N}^{+}, N=7$), are initially embedded. These embeddings are then concatenated to form a  lane feature:
\begin{equation}
\mathbf{F}_l =\operatorname{Embed}(  \operatorname{Embed}(s_{l,1})\oplus \ldots \operatorname{Embed}(s_{l,7})),
\end{equation}
where symbol $\oplus$ denotes the concatenation operation, and there are a total of 520 ($(1\times4+4)\times 7+ (28\times16+16)$) parameters. 
Multi-head self-attention (MHA)~\cite{attention} is  employed to integrate features from all incoming lanes relevant to each phase:
\begin{equation}
\mathbf{F}^p = \operatorname{Mean}(\operatorname{MHA}(\mathbf{F}_l\oplus \mathbf{F}_k), l, k\in \mathcal{L}_p,
\end{equation}
where $p$ denotes each phase, $\mathcal{L}_p$ is the set of participating entering lanes of phase $p$. We utilize a self-attention mechanism with 4 heads, each having a key (query and value) dimension of 16, resulting in a total of 4,304 parameters, which is calculated as $4 \times 3 \times (16 \times 16 + 16) + 4 \times 16 \times 16 + 16$.
\item \textbf{Q-value prediction for phase control.} The Fused phase feature is processed using MHA to capture inter-phase correlations:
    \begin{equation}
        \mathbf{F}^{\mathcal{P}} = \operatorname{MHA}(\mathbf{F}^{p1} \oplus \mathbf{F}^{p2} \oplus \mathbf{F}^{p3} \oplus \mathbf{F}^{p4} ), \mathcal{P} = \{p1,p2,p3,p4\},
    \end{equation}
 We utilize a self-attention mechanism with 4 heads, each having a key (query and value) dimension of 8, resulting in a total of 2160 parameters.
    The correlated phase features are further embedded to derive the Q-values:
    \begin{equation}
        \widetilde q_p = \operatorname{Embed}(\mathbf{F}^{\mathcal{P}}).
    \end{equation}
    We utilize 3 MLP layers, and there are 1201 (i.e., $(16\times20+20)+(20\times20+20)+(20\times1+1)$) parameters.
    The phase action is selected based on the maximum Q-value, denoted as $p^{\star}$.
\item \textbf{Q-value prediction for duration control.} The determined phase $p^{\star}$ is represented as a matrix ${\bm H}$, which is utilized to extract features specific to $p^{\star}$:
\begin{equation}
\mathbf{F}^{p{\star}} = {\bm H} \times (\mathbf{F}^{p1} \oplus \mathbf{F}^{p2} \oplus \mathbf{F}^{p3} \oplus \mathbf{F}^{p4} ),
\end{equation}
and there are no parameters.
The phase features are then embedded to calculate the Q-value of each duration action:
\begin{equation}
\widetilde q_d = \operatorname{DuelingBlock}(\operatorname{Embed}(\mathbf{F}^{p^{\star}})),
\end{equation}
the $\operatorname{DuelingBlock}$ is a specialized structure as described in~\cite{dueling}. 
There are two MLPs and one dueling block, and a total of 1768 (i.e., $(16\times20+20)+(20\times20+20)\times3+(20\times1+1)+(20\times7+7)$) parameters.
The duration action is determined with the maximum Q-value.
\end{itemize}
In summary, DynamicLight contains 9533 parameters.

\subsection{Ablation Study on DynamicLight Training Techniques}
\label{apxsec:training}
In training our DynamicLight model, we employ a variety of techniques, each contributing uniquely to the training process: \textbf{fix $\theta_1$} means fix $\theta_1$ when training duration control;
\textbf{clear memory1} means clear the replay memory of phase control before training duration control;
\textbf{clear memory2} means clear the replay memory of duration control before fine-tuning phase control and duration control together;
\textbf{reset lr1} means to reset a lower learning rate of duration control after 30 steps;
\textbf{reset lr2} means to reset a lower learning rate of fine-tuning phase control and duration control;
\textbf{train seperately} means first train phase control and next train duration control;
and \textbf{fine-tuning} means train phase control and duration control together with a lower learning rate.
In the separate training process of DynamicLight, the duration action is consistently fixed at 15 seconds to facilitate the learning of phase control. Concurrently, the optimal phase action is employed to enhance the effectiveness of duration control. This approach ensures a focused and efficient training strategy for each control aspect.

Table~\ref{apx:training} displays a comparative analysis of DynamicLight's performance with and without the implementation of these training techniques, demonstrating the effectiveness of the approaches we have employed.

\begin{table}[H]
\caption{Performance comparison of DynamicLigh with and without some training techniques (ATT in seconds).}
\label{apx:training}
    \centering 
    \begin{tabular}{lccc|cc}
    \toprule
    \multirow{2}{*}[-0.5em]{Trainig config}&\multicolumn{3}{c}{JN dataset}&\multicolumn{2}{c}{HZ dataset}\\\cmidrule{2-6}&JN1&JN2&JN3&HZ1&HZ2\\\midrule
    w/o fix $\theta_1$ & $240.41 $& $221.98 $& $219.55 $& $264.33 $& $ 314.87$\\
    w/o clear memory1 & $ 240.06$& $227.86 $& $222.79 $& $ 264.55$& $319.22 $\\
    w/o clear memory2 & $237.16 $& $221.93 $& $220.52 $& $\mathbf{261.93} $& $310.23 $\\
    w/o reset lr1 & $ 237.34$& $ 221.64$& $ 219.94$& $263.19 $& $313.36 $\\
    w/o reset lr2 & $ 236.40$& $221.64 $& $\mathbf{218.22} $& $ 262.94$& $314.63 $\\
    w/o train separately & $241.29 $& $224.77 $& $223.16 $& $263.43 $& $307.51 $\\
    w/o fine-tuning & $ 239.16$& $224.80 $& $220.93 $& $ 263.60$& $309.99 $\\
    \midrule
    DynamicLight & $\mathbf{235.95} $& $\mathbf{221.29} $& $218.66 $& $262.24 $& $\mathbf{303.23} $\\
    \bottomrule
    \end{tabular}
\end{table}
\subsection{Impact of Fixed Duration Actions}
\label{apxsec:fixd}
When training DynamicLight separately, the selection of a predetermined duration action plays a crucial role in shaping the phase control performance of DynamicLight, as highlighted in previous studies~\cite{advanced, ql}. To understand this impact in depth, we conducted evaluations of DynamicLight's performance using different fixed duration actions. Specifically, we chose 10, 15, and 20 seconds as the fixed duration actions from the set $\{10, 15, 20, 25, 30, 35, 40\}$.

Table~\ref{apxsec:fixd} presents the results of DynamicLight under these varying fixed duration actions. Based on our analysis, we have determined that a fixed duration of 15 seconds optimizes performance, and thus, we have selected it as the default duration action for our model.

\begin{table}[H]
\caption{Performance comparison of DynamicLigh under different fixed duration actions (ATT in seconds).}
\label{apx:fixd}
    \centering 
    \begin{tabular}{cccc|cc}
    \toprule
    \multirow{2}{*}[-0.5em]{Fixed duration action}&\multicolumn{3}{c}{JN dataset}&\multicolumn{2}{c}{HZ dataset}\\\cmidrule{2-6}&JN1&JN2&JN3&HZ1&HZ2\\\midrule
    10s & $ 239.73$& $222.24 $& $220.01 $& $ 267.51$& $318.93 $\\
    15s &$\mathbf{235.95} $& $\mathbf{221.29} $& $\mathbf{218.66} $& $\mathbf{262.24} $& $\mathbf{303.23} $\\
    20s & $ 239.63$& $ 223.59$& $ 219.95$& $ 263.90$& $311.56 $\\
    \bottomrule
    \end{tabular}
\end{table}

\section{Details of Experimental Results}
\label{apx:exp_results}

\subsection{DynamicLight Variants' Network Details}
\label{apxsec:variants}
All variants of the DynamicLight model utilize the number of vehicles under the segmented road (four segments of 100 meters each) as the state representation for duration control. This approach is consistent across all variants, maintaining the same network structure for duration control as in the original DynamicLight. 
\begin{itemize}[leftmargin=*]
\item \textbf{Lane feature embedding.} The characteristics of each entering lane are initially embedded. These embeddings are then concatenated to form a comprehensive lane feature:
\begin{equation}
\mathbf{F}_l = \operatorname{Embed}( \operatorname{Embed}(s_{l,4})\oplus \ldots \operatorname{Embed}(s_{l,7})),
\end{equation}
where symbol $\oplus$ denotes the concatenation operation. There are 5 MLPs and a total of 304 (i.e., $(1\time4 +4)\times4 +(16\times16+16)$) parameters.
\item \textbf{Phase feature construction.} 
Multi-head self-attention (MHA) is employed to integrate features from all incoming lanes relevant to each phase:
\begin{equation}
\mathbf{F}^p = \operatorname{Mean}(\operatorname{MHA}(\mathbf{F}_l\oplus \mathbf{F}_k), l, k\in \mathcal{L}_p,
\end{equation}
where $p$ represents one phase, and $\mathcal{L}_p$ is the set of participating entering lanes of phase $p$.  We utilize a self-attention mechanism with 4 heads, each having a key (query and value) dimension of 16, resulting in a total of 4,304 parameters, which is calculated as $4 \times 3 \times (16 \times 16 +  16) + 4 \times 16 \times 16 + 16$.
\item \textbf{Phase feature selection.} The phase $p^{\star}$ is determined by the phase control strategy. The corresponding feature of  $p^{\star}$ is extracted 
\begin{equation}
\mathbf{F}^{p{\star}} = {\bm H} \times (\mathbf{F}^{p1} \oplus \mathbf{F}^{p2} \oplus \mathbf{F}^{p3} \oplus \mathbf{F}^{p4} ),
\end{equation}
where $H$ is a matrix representing $p^{\star}$ for feature selection by matrix multiplication.
\item \textbf{Q-value prediction for duration control.} The phase features are further processed to obtain the Q-value of each duration action:
\begin{equation}
\widetilde q_d = \operatorname{DuelingBlock}(\operatorname{Embed}(\mathbf{F}^{p^{\star}})).
\end{equation}
\end{itemize}
There are 2 MLPs and one dueling block, and a total of 1768 ($(16\times20+20)+(20\times20)\times3+(20\times1+1)+(20\times7+7)$) parameters.
Finally, the duration action with the maximum Q-value is selected.
This network contains 6376 parameters.
\subsection{DynamicLight-Lite's Network Design}
\label{apxsec:lite}
Based on a comprehensive understanding and analysis of the DynamicLight variants network, we introduce DynamicLight-Lite.  Compared to DynamicLight variants, the DQN of DynamicLight-Lite for duration control is different from the following:
\begin{itemize}
\item \textbf{Lane feature embedding.} DynamicLight-Lite embeds all the features of lane $l$ into a one-dimensional latent space with an MLP to obtain each lane's feature:
\begin{align}
    s_l & = s_{l,4}\oplus \ldots \oplus s_{l,7}, \\
  \mathbf{ F}_l & = \sigma(s_l{\bm W}_e+{\bm b}_e),
    \end{align}
where $\oplus$ is the concatenate operation , $s_l\in\mathbb{R}^{4}$, ${\bm W}_e\in\mathbb{R}^{4\times 1}$ and ${\bm b}_e\in\mathbb{R}^{1}$ are weight matrix and bias vector to learn, and $\sigma$ is the sigmoid function. The number of trainable parameters in this part is 5 ($4\times1+1$).
\item \textbf{Phase feature construction.}  The features of $l, l\in\mathcal{L}_p$ are directly added to derive the feature of phase $p$. 
\begin{equation}
    \mathbf{F}^p = \operatorname{Add}(\mathbf{F}^l\oplus \mathbf{F}^k), l, k \in \mathcal{L}_p.
\end{equation}
There are no trainable parameters in this part. Refer to Appendix~\ref{apx:fusion} for more details on the performance of this feature fusion method.
\item \textbf{Phase feature selection.} The phase $p^{\star}$ is determined by the phase control strategy. The corresponding feature of  $p^{\star}$ is extracted by
\begin{equation}
\mathbf{F}^{p{\star}} = {\bm H} \times (\mathbf{F}^{p1} \oplus \mathbf{F}^{p2} \oplus \mathbf{F}^{p3} \oplus \mathbf{F}^{p4} ),
\end{equation}
where $H$ is a matrix representing $p^{\star}$ for feature selection by matrix multiplication.
There are no trainable parameters either.
\item \textbf{Q-value prediction.} An MLP is directly used to obtain the Q-value of each duration action:
    \begin{equation}
        \widetilde{q}_d = \operatorname{Embed}(\mathbf{F}^{p\star}) = \sigma(\mathbf{F}^{p\star}{\bm W}_c+{\bm b}_c),
    \end{equation}
where ${\bm W}_c\in\mathbb{R}^{1\times 7}$ and ${\bm b}_c\in\mathbb{R}^{7}$ are weight matrix and bias vector to learn, $\sigma$ is the ReLU function. The trainable parameters in this part are 14 (i.e., $1\times7+7$).

\end{itemize}

In summary, DynamicLight-Lite employs MLP twice: the first MLP contains 5 trainable parameters, and the last MLP contains 14 trainable parameters. Other modules do not contain trainable parameters, bringing the total count to 19. With significantly fewer parameters than other RL models for TSC, DynamicLight-Lite is highly efficient. Table~\ref{apx:clite} compares different models in terms of parameter counts, memory usage, and inference times (for 1000 random samples). The reported memory usage and inference time are averages of five independent results. When built using TensorFlow2, DynamicLight-Lite exhibits the lowest memory and inference time. Its minimal parameter count of just 19 also makes it highly feasible to implement with Numpy, further reducing memory usage and inference time.
    
\begin{table}[H]
\caption{ Comparison of different models by their parameter counts, memory usage, and inference times.}
\label{apx:clite}
\centering 
\begin{threeparttable}
\begin{tabular}{lcccc}\toprule
Model & Parameters & Memory usage (KB)& Inference time (Second) \\\midrule
FRAP~\cite{frap} & $1369$& $12961$& $0.31$ \\
CoLight~\cite{colight}&$18756$ &$12732$ & $0.32$\\
AttentionLight~\cite{ql} &$3101$ & $13466$& $0.33$ \\
DynamicLight-MQL &$6376$ & $13662$& $0.32$ \\
DynamicLight & $9533$&$14232$ &$0.39$ \\
DynamicLight-Lite &$\mathbf{19}$ & $12342$& $0.20$\\
DynamicLight-Lite (Numpy) &$\mathbf{19}$ & $-$&$\mathbf{4.6}\mathrm{e}{-4}$\\\bottomrule
\end{tabular}
\begin{tablenotes}
\footnotesize
\item[$\star$] DynamicLight-Lite (Numpy) is realized with Numpy and the memory usage is too small to record.
\end{tablenotes}
\end{threeparttable}
\end{table}

\subsection{Effects of Duration Control}
\begin{table}[ht]
\setlength\tabcolsep{16pt}
\caption{Performance comparisons of DynamicLight and its variants with and without the duration control strategy on real-world datasets (ATT in seconds).}
\label{tab:apx_ablation}
\centering
\begin{tabular}{lccc|cc}\toprule
\multirow{2}{*}[-0.5em]{Model}&\multicolumn{3}{c}{JN dataset}&\multicolumn{2}{c}{HZ dataset}\\\cmidrule{2-6}&JN1&JN2&JN3&HZ1&HZ2\\\midrule

w/o duration&$244.2$8&$231.82$&$227.57$&$270.29$&$312.76$\\
DynamicLight&$\mathbf{235.95}$&$\mathbf{221.29}$&$\mathbf{218.66}$&$\mathbf{262.24}$&$\mathbf{303.23}$\\
Improvement&$\uparrow$ $3.41$\%&$\uparrow$ $4.54$\%&$\uparrow$ $3.92$\%&$\uparrow$ $2.98$\%&$\uparrow$ $3.05$\%\\\midrule

w/o duration&$559.49$&$506.72$&$513.10$&$560.08$&$466.66$\\
DynamicLight-Rand&$\mathbf{348.48}$ &$\mathbf{322.78}$ &$\mathbf{319.23}$ &$\mathbf{372.91}$ &$\mathbf{386.50}$ \\
Improvement&$\uparrow$ $37.71$\%&$\uparrow$ $36.03$\%&$\uparrow$ $37.78$\%&$\uparrow$ $33.42$\%&$\uparrow$ $17.18$\%\\\midrule

w/o duration&$429.27$&$370.34$&$384.89$&$497.87$&$408.31$\\
DynamicLight-FT&$\mathbf{279.99}$ &$\mathbf{281.87}$ &$\mathbf{274.93}$ &$\mathbf{320.49}$ &$\mathbf{327.43}$ \\
Improvement&$\uparrow$ $34.78$\%&$\uparrow$ $23.89$\%&$\uparrow$ $28.57$\%&$\uparrow$ $35.63$\%&$\uparrow$ $19.81$\%\\\midrule

w/o duration&$274.99$&$246.41$&$244.63$&$289.55$&$349.85$\\
DynamicLight-MP&$\mathbf{253.52}$ &$\mathbf{240.77}$ &$\mathbf{235.64}$ &$\mathbf{281.45}$ &$\mathbf{328.53}$ \\
Improvement&$\uparrow$ $7.81$\%&$\uparrow$ $2.29$\%&$\uparrow$ $3.67$\%&$\uparrow$ $2.80$\%&$\uparrow$ $6.09$\%\\\midrule

w/o duration&$268.87$&$240.02$&$238.51$&$284.32$&$325.44$\\
DynamicLight-MQL&$\mathbf{241.56}$ &$\mathbf{230.63}$ &$\mathbf{225.31}$ &$\mathbf{272.19}$ &$\mathbf{303.92}$ \\
Improvement&$\uparrow$ 10.16\%&$\uparrow$ 3.91\%&$\uparrow$ 5.53\%&$\uparrow$ 4.27\%&$\uparrow$ 6.61\%\\\midrule

w/o duration&$268.87$&$240.02$&$238.51$&$284.32$&$325.44$\\
DynamicLight-Lite&$\mathbf{249.98}$ &$\mathbf{239.48}$ &$\mathbf{230.92}$ &$\mathbf{283.85}$ &$\mathbf{305.78}$ \\
Improvement&$\uparrow$ 7.03\%&$\uparrow$ 0.22\%&$\uparrow$ 3.18\%&$\uparrow$ 0.17\%&$\uparrow$ 6.04\%\\\bottomrule
\end{tabular}
\end{table}

\subsection{Learning Capability Evaluation}
\begin{table}[ht]
\setlength\tabcolsep{16pt}
\caption{Learning capability of DynamicLight and its variants comparison with maintaining their duration control strategy and changing their phase control strategy as FixedTime (ATT in seconds).}
\label{tab:apx_cap}
\centering
\begin{tabular}{lccc|cc}\toprule
\multirow{2}{*}[-0.5em]{Model}&\multicolumn{3}{c}{JN dataset}&\multicolumn{2}{c}{HZ dataset}\\\cmidrule{2-6}&JN1&JN2&JN3&HZ1&HZ2\\\midrule

DynamicLight&$\mathbf{263.90}$&$\mathbf{245.24}$&$\mathbf{241.00}$&$328.31$&$342.77$\\
DynamicLight-Rand&$301.82$&$277.65$&$278.01$&$322.79$&$344.96$\\
DynamicLight-FT&$279.99$&$281.87$&$274.93$&$320.49$&$327.43$\\
DynamicLight-MP&$283.80$&$264.45$&$272.25$&$310.69$&$352.27$\\
DynamicLight-MQL&$270.32$&$250.09$&$245.96$&$\mathbf{309.14}$&$\mathbf{ 319.78}$\\
DynamicLight-Lite&$277.61$&$254.02$&$269.32$&$506.53$&$324.72$\\\bottomrule
\end{tabular}
\end{table}

\subsection{Performance of DynamicLight on Synthetic Datasets}
\begin{table}[ht]
\setlength\tabcolsep{8pt}
\caption{Performance comparisons of DynamicLight and its variants on the synthetic datasets (ATT in seconds).}
\label{tab:apx_syn}
\centering
\begin{tabular}{lcc|cc|cc}\toprule
\multirow{2}{*}[-0.5em]{Model}&\multicolumn{2}{c}{Topology 1}&\multicolumn{2}{c}{Topology 2}&\multicolumn{2}{c}{Topology 3}\\\cmidrule{2-7}&SYN1-1&SYN1-2&SYN2-1&SYN2-2&SYN3-1&SYN3-2\\\midrule
FixedTime~\cite{fixedtime}&$448.47$&$623.64$&$360.86$&$538.81$&$289.63$&$330.10$\\
Max Pressure~\cite{mp2013}&$249.19$&$349.35$&$183.86$&$238.07$&$337.67$&$420.30$\\
Max QueueLength~\cite{ql}&$180.61$&$223.00$&$168.10$&$197.38$&$272.18$&$283.30$\\\midrule

CoLight~\cite{colight}&$197.86$&$294.38$&$182.48$&$228.61$&$229.52$&$283.30$\\
Advanced-CoLight~\cite{advanced}&$197.28$&$295.83$&$170.96$&$234.69$&$216.87$&$309.77$\\\midrule

DynamicLight-Rand&$357.14$&$476.96$&$295.35$&$367.48$&$225.90$&$242.92$\\
DynamicLight-FT&$254.75$&$292.10$&$222.33$&$255.45$&$169.58$&$187.25$\\
DynamicLight-MP&$214.65$&$249.75$&$178.58$&$208.04$&$160.24$&$166.90$\\
DynamicLight-MQL&$170.31$&$199.86$&$157.26$&$181.30$&$149.81$&$159.53$\\
DynamicLight-Lite&$175.25$&$212.94$&$162.68$&$187.56$&$157.99$&$172.47$\\\midrule
DynamicLight&$\mathbf{163.64}$&$\mathbf{195.20}$&$\mathbf{150.29}$&$\mathbf{174.49}$&$\mathbf{142.16}$&$\mathbf{153.79}$\\
Improvement&$\uparrow$ $9.40$\%&$\uparrow$ $12.47$\%&$\uparrow$ $10.59$\%&$\uparrow$ $11.6$\%&$\uparrow$ $34.45$\% &$\uparrow$ $45.71$\%\\
\bottomrule
\end{tabular}
\end{table}




\end{appendices}
\end{document}